\documentclass{article}
\usepackage{fullpage}

\usepackage{longtable}
\usepackage{indentfirst}
\usepackage{comment}
\usepackage{lineno,hyperref}
\usepackage{longtable}
\modulolinenumbers[5]
\usepackage{graphicx}
\usepackage{amsfonts}
\usepackage{verbatim}
\usepackage{amsmath}
\usepackage{algorithm}
\usepackage{caption}
\usepackage{algorithmicx}
\usepackage{algpseudocode}
\floatname{algorithm}{Algorithm}

\usepackage{microtype}
\usepackage{graphicx}
\usepackage{subfigure}
\usepackage{bm}
\usepackage{epstopdf}

\usepackage{hyperref}
\usepackage{authblk}
\usepackage{booktabs}
\usepackage{makecell}
\allowdisplaybreaks

\author{Jing Zhao, Jingjing Fei, Shiliang Sun}
\affil{East China Normal University}
\title{A Variant of Gaussian Process Dynamical Systems}
\begin{document}
\date{November 7, 2018}
\maketitle
\section{Abstract}
In order to better model high-dimensional sequential data, we propose a collaborative multi-output Gaussian process dynamical system (CGPDS), which is a novel variant of GPDSs. The proposed model assumes that the output on each dimension is controlled by a shared global latent process and a private local latent process. Thus, the dependence among different dimensions of the sequences can be captured, and the unique characteristics of each dimension of the sequences can be maintained. For training models and making prediction, we introduce inducing points and adopt stochastic variational inference methods.
\section{Introduction}
Sequential data are very common in the real world, such as human activities, meteorological data, and video clips. Particularly, some sequential data are high dimensional. For example, most of the videos are high resolution and may have the megapixel resolution. Modeling high-dimensional sequential data where the number of dimension $D$ is much larger than the number of points $N$ (i.e., $D \gg N$) is significant and challenging. Dynamical systems are often used for modeling sequential data. Among these, Gaussian process dynamical systems (GPDSs) are the recently proposed probabilistic models and have achieved excellent results in several applications on sequential data. Existing GPDSs have their own advantages in different aspects, but they also have some limitations when dealing with complex high-dimensional sequential data. We aim to develop a novel GPDS to model high-dimensional sequential data more rationally.
\par
GPDSs employ Gaussian processes (GPs) to model the dynamics and nonlinear mappings in the systems. GPs are stochastic processes over real-valued functions and defined by mean functions and covariance functions \cite{GPML}. The standard GPs are often used as regression or classification models. For dealing with multi-task or multi-output problems, some multi-output GPs are developed such as the multi-task GP \cite{Bronilla:multitask}, convolved multi-output GP \cite{Alvarez09,Alvarez:computationally}, GP regression network \cite{Wilson:gaussian}, collaborative multi-output Gaussian process (COGP) \cite{COGP}, warped multi-output Gaussian process \cite{kaiser2018bayesian} and heterogeneous multi-output Gaussian process \cite{moreno2018heterogeneous}. GP latent variable models (GPLVMs) were proposed to implement nonlinear dimensionality reduction for high-dimensional data, which employed the shared latent variables and assumed the conditionally independence for multiple outputs \cite{GPLVM04,GPLVM05,BGPLVM,atkinson2018structured}. GPLVMs also provide inspirations for the subsequent research on multi-dimensional sequential models.
\par
GPDSs extended the GPLVM by adding specific dynamical priors on the latent variables. For example, GP dynamical models (GPDMs) \cite{GPDM}, variational GPDSs (VGPDSs) \cite{VGPDS} and variational dependent multi-output GPDSs (VDM-GPDSs) \cite{VDM-GPDS} are the state-of-the-art GPDSs. The GPDM models the dynamics by adding a Markov dynamical prior on the latent space, and characterizes the variability of outputs by constructing the variance of outputs with different parameters. GPDMs were applied in some practical applications, such as object tracking \cite{PeopleTracking}, computer animation \cite{Henter} and activity recognition \cite{minguez2018pedestrian}. The VGPDS imposes a GP dynamical prior to the latent space of GPLVM, which can capture some specific dynamics such as periodicity by periodic kernels. VGPDSs were also applied in many fields, such as phoneme classification \cite{PhonemeVGPDS}, video repairing \cite{xiong2016dual} and multi-task motion modeling \cite{Korkinof2017multitask}. The VDM-GPDS considers the dependence of multiple outputs and introduces convolution processes to explicitly depict multi-output dependence. The VDM-GPDS achieved better performance in some applications such as sequence forecast and sequence recovery than GPDMs and VGPDSs, but took a long time for training \cite{VDM-GPDS}.

\par
The existing GPDSs mentioned above have some limitations when modeling high-dimensional sequential data. For example, GPDMs and VPGDSs ignore the dependence and differences among multiple outputs. VDM-GPDSs contains overly complex structures and is difficult to process high-dimensional data. In this paper, we propose a novel collaborative GPDS (CGPDS) for modeling high-dimensional sequential data. In the CGPDS, each output is constructed by a global process and a local latent process. The global latent process and the local latent process are used to capture the universality and individuality of the output. We adopt the variational Bayesian inference to our model, which would avoid overfitting. Furthermore, the outputs are conditionally independent and the resulting evidence lower bound can be decomposed across dimensions, which enables the model to handle high-dimensional sequential data.

\section{The Collaborative Multi-output GPDS}
Multi-output sequential data are often denoted as $\{\mathbf y_n,t_n\}_{n=1}^N$, where $\mathbf y_n \in \mathbb R^D$ is an observation at time $t_n \in \mathbb R^+$. We assume that there are some low-dimensional latent variables $\mathbf x_n \in \mathbb R^Q$ (with $Q \ll D$) that govern the generation of the observed outputs. The low-dimensional latent variables are assumed to have a GP prior which is used to model the mapping from $t_n$ to $\mathbf{x}_n$, as in \cite{VGPDS}. Particularly, we use the global latent processes $\mathbf{h}$ and local latent processes $\bm{\ell}_d$ to construct the observed outputs $\mathbf{y}$.

\begin{figure}
  \centering
  \includegraphics[width=0.3\textwidth]{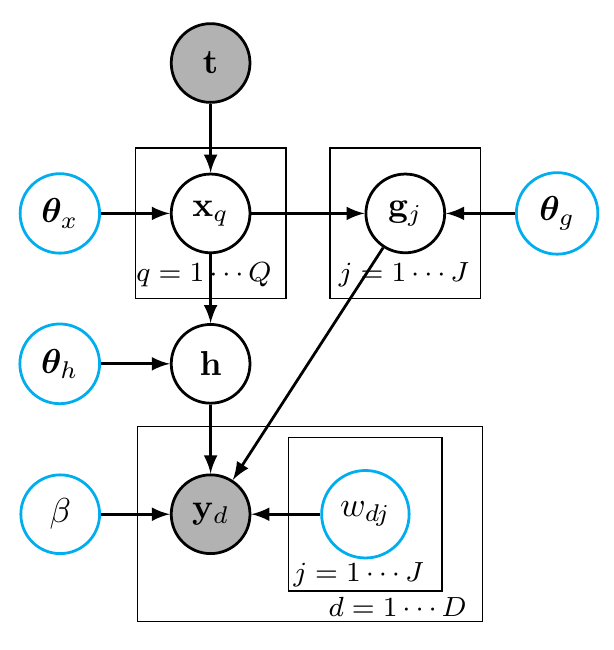}\\
  \caption{The graphical model for the CGPDS. The gray solid circles represent observations. The black hollow circles represent latent variables. The cyan hollow circles represent parameters.}\label{fig:model}
\end{figure}

Specifically, the proposed CGPDS is composed of four layers which built three mappings. They are an independent multi-output GP mapping from the time indices $\mathbf t$ to the low-dimensional latent space $X$, independent multi-output GP mappings from $X$ to the latent spaces $\mathbf h$ and $\{\mathbf g_j\}_{j=1}^J$, and a linear Gaussian mapping from $\mathbf h$ and $\{\bm\ell_d\}_{d=1}^D$ to the observation space $Y$. The graphical model for the CGPDS is shown in Figure \ref{fig:model}.

In the first mapping, the low-dimensional latent variable $X$ is assumed to be a multi-output GP indexed by time $t$, i.e.,
\begin{equation}
x_q(t) \sim \mathcal GP(0,{\kappa}_x(t,t')),q = 1,\ldots,Q,
\end{equation}
where individual dimensions of the latent function $\mathbf x(t)$ are independent sample paths drawn from a GP with the covariance function $\kappa_x(t,t')$ whose parameters are $\bm \theta_x$.

In the second mapping, both latent variables $\mathbf h$ and $\{\mathbf g_j\}_{j=1}^J$ are multi-output GPs with the input $\mathbf x$, and
\begin{eqnarray}
  {h}(\mathbf x) &\sim& \mathcal GP(0,\kappa_h(\mathbf x,\mathbf x')),\\
  {g_j}(\mathbf x) &\sim& \mathcal GP(0,\kappa_g^j(\mathbf x,\mathbf x')).
\end{eqnarray}
Here, the covariance functions $\kappa_h(\mathbf x,\mathbf x')$ and $\kappa_g^j(\mathbf x,\mathbf x')$ are parameterized by $\bm \theta_h$ and $\bm \theta_g^j$, respectively.

In the third mapping, we first construct the local latent variables $\bm\ell_d$ as a weighted average of latent processes $\{\mathbf g_j\}_{j=1}^J$, i.e., $\bm\ell_d=\sum_{j=1}^{J}w_{dj}\mathbf{g}_j$. The different weights $\{w_{dj}\}$ represent the local parameters for $D$ outputs. Therefore, the resulting ${\bm \ell}_d$ is unique to the $d$th output $\mathbf y_d$. Then $\mathbf y_d$ is assumed to be the sum of a local latent process ${\bm {\ell}}_d$ and a global latent process $\mathbf h$ which is shared for $D$ dimensions. The likelihood with standard iid Gaussian noise is given by
\begin{eqnarray}\label{eq:yd}
p(\mathbf y_d \vert \mathbf g,\mathbf h) &=& \mathcal{N}(\mathbf y_d; {\bm{\ell}}_d + \mathbf h,\beta ^{-1}I)\nonumber\\
&=&\mathcal N(\mathbf y_d;\sum_{j = 1}^J {w_{dj} \mathbf g_j + \mathbf h} ,\beta ^{-1}I),
\end{eqnarray}
where $\beta$ is the inverse variance of the white Gaussian noise.
Therefore, we can extract the dependence among the dimensions through the global latent process $\mathbf h$ and maintain the characteristics of each dimension through the local latent process $\bm\ell_d$.

\section{Variational Inference}
The variational inference is adopted to our model, which requires maximizing the variational lower bound of the logarithmic marginal likelihood.

First, we introduce inducing variables $\{\mathbf u_j\}_{j=1}^J$ and $\mathbf v$, where $\mathbf u_j$ represents the value of $g_j(\mathbf x)$ at the inducing inputs $Z_g^j$ and $\mathbf v$ represents the value of $h(\mathbf x)$ at the inducing inputs $Z_h$. We assume that all the latent processes have the same number of inducing points, $M$. According to the independence of processes $\{\mathbf g_j\}_{j=1}^J$, we have
\begin{eqnarray}
p(\mathbf v) &= &\mathcal N(\mathbf v;0,\mathbf K_{\mathbf v,\mathbf v}),\\
p(\mathbf h|\mathbf v) &=& \mathcal N\left(\mathbf h;\mathbf K_{\mathbf h,\mathbf v} (\mathbf K_{\mathbf v,\mathbf v})^{-1}\mathbf v,\mathbf K_{\mathbf h,\mathbf h}-\mathbf K_{\mathbf h,\mathbf v}(\mathbf K_{\mathbf v,\mathbf v})^{-1}\mathbf K_{\mathbf v,\mathbf h}\right),\\
    p(\mathbf u) &=& \prod_{j=1}^{J}\mathcal N(\mathbf u_j;0,\mathbf K_{\mathbf u,\mathbf u}^j ),\\
  p(\mathbf g|\mathbf u) &=& \prod_{j=1}^{J}\mathcal N\left(\mathbf g_j;\mathbf K_{\mathbf g,\mathbf u}^j (\mathbf K_{\mathbf u,\mathbf u}^j)^{-1}\mathbf u_j,\mathbf K_{\mathbf g,\mathbf g}^j-\mathbf K_{\mathbf g,\mathbf u}^j(\mathbf K_{\mathbf u,\mathbf u}^j)^{-1}\mathbf K_{\mathbf u,\mathbf g}^j\right),
  \end{eqnarray}
where $\mathbf u = \{\mathbf u_j\}_{j=1}^J$, $\mathbf K_{\mathbf v,\mathbf v} = \kappa_h(Z_h,Z_h)$, $\mathbf K_{\mathbf h,\mathbf v} = \kappa_h(X,Z_h)$, $\mathbf K_{\mathbf h,\mathbf h} = \kappa_h(X,X)$ and $\mathbf K_{\mathbf v,\mathbf h} = \mathbf K_{\mathbf h,\mathbf v}^\top$. Similarly, $\mathbf K_{\mathbf u,\mathbf u}^j = \kappa_g^j(Z_g^j,Z_g^j)$, $\mathbf K_{\mathbf g,\mathbf u}^j = \kappa_g^j(X,Z_g^j)$, $\mathbf K_{\mathbf g,\mathbf g}^j = \kappa_g^j(X,X)$ and $\mathbf K_{\mathbf u,\mathbf g}^j = (\mathbf K_{\mathbf g,\mathbf u}^j)^\top$.

Then, we assume that the variational distribution has the following form:
\begin{equation}
q(\mathbf g,\mathbf h,\mathbf u,\mathbf v,X|\mathbf y,\mathbf t) =
p(\mathbf g|\mathbf u)p(\mathbf h|\mathbf v)q(\mathbf u,\mathbf v)q(X).
\end{equation}
Since the conditional distributions $p(\mathbf g|\mathbf u)$ and $p(\mathbf h|\mathbf v)$ are known, we only need to learn $q(\mathbf u,\mathbf v)q(X)$ by minimizing the KL divergence between the approximate posterior and true posterior.

Finally, we can derive the evidence lower bound (ELBO) of the log marginal likelihood,
\begin{eqnarray}
\mathcal{L}&=&\int q(\mathbf u,\mathbf v)q(X)\log{\frac{p(\mathbf y|\mathbf u,\mathbf v,X)p(\mathbf u,\mathbf v)}{q(\mathbf u,\mathbf v)}}d\mathbf ud\mathbf vdX-\sum_{q=1}^Q{\mathrm{KL}[q(\mathbf x_q)||p(\mathbf x_q)]},
\end{eqnarray}
which can be optimized by gradient-based approaches.

The obtained ELBO can be decomposed across $D$ outputs. This decomposition makes it possible to apply the stochastic variational inference, thus allowing the proposed model to handle high-dimensional sequential data.
\section{Prediction}
Our model is capable of performing prediction for high-dimensional sequential data in two situations. One is prediction with time which uses only time to generate completely new sequences. The other one is prediction with time and partial observations, which can be seen as reconstructing missing observations using partial observations.
\subsection{Generation}
For generation, we need to compute the posterior $p(Y_*|Y)$, where $Y_* \in \mathbb R^{N_*\times D}$ represents the predicted outputs,
\begin{eqnarray}
 p(Y_*|Y)&=&\int p(Y_*|\mathbf g_*,\mathbf h_*) p(\mathbf g_*,\mathbf h_*|Y)d\mathbf g_*d\mathbf h_*,\\\nonumber
&=&\int p(Y_*|\mathbf g_*,\mathbf h_*)p(\mathbf g_*,\mathbf h_*|X_*,Y)p(X_*|Y)d\mathbf g_*d\mathbf h_*dX_*,
\end{eqnarray}
where $\mathbf g_* \in \mathbb R^{N_* \times J}$ and $\mathbf h_* \in \mathbb R^{N_*}$ denote the set of latent variables for the predicted outputs $Y_*$, and $X_* \in \mathbb R^{N_* \times Q}$ represents the corresponding low-dimensional latent variables.
The distribution $p(\mathbf g_*,\mathbf h_*|X_*,Y)$ is approximated by
\begin{eqnarray}
  p(\mathbf g_*,\mathbf h_*|X_*,Y)  \approx q(\mathbf g_*,\mathbf h_*|X_*)=\int p(\mathbf g_*,\mathbf h_*|\mathbf u,\mathbf v,X_*)q(\mathbf u,\mathbf v)d\mathbf u d\mathbf v .\nonumber
\end{eqnarray}

Since $p(\mathbf g_*,\mathbf h_*|\mathbf u,\mathbf v,X_*)$ and the optimal $q(\mathbf u,\mathbf v)$ in our variational framework are both Gaussian, the approximate distribution $ q(\mathbf g_*,\mathbf h_*|X_*)$ is also Gaussian and can be computed analytically. As the distribution $p(X_*|Y)$ is approximated by the variational $q(X_*)$, the joint posterior density of $\mathbf g_*$ and $\mathbf h_*$ can be obtained by
\begin{equation}
  p(\mathbf g_*,\mathbf h_*|Y)\approx\int q(\mathbf g_*,\mathbf h_*|X_*)q(X_*)dX_*.
\end{equation}

Although the integration of $q(\mathbf g_*,\mathbf h_*|X_*)$ w.r.t $q(X_*)$ is not analytically feasible, we can calculate the expectation of $\mathbf g_*$ and $\mathbf h_*$ by following \cite{VGPDS}, which are denoted as $\mathbb{E}(\mathbf g_*)$ and $\mathbb{E}(\mathbf h_*)$, respectively. The element-wise autocovariance matrices of $\mathbf g_*$ and $\mathbf h_*$  are represented as $\mathbb{V}(\mathbf g_*)$ and $\mathbb{V}(\mathbf h_*)$, respectively. Since $Y_{*d}=\sum_{j=1}^{J} w_{dj}\mathbf g_{*j}+\mathbf h_{*}, d \in [1 \dots D]$, the expectation and covariance of $Y_{*d}$ are $\mathbb{E}(Y_{*d})=\sum_{j=1}^{J} w_{dj}\mathbb{E}(\mathbf g_{*j})+E(\mathbf h_{*})$ and $\mathbb{V}(Y_{*d})=\sum_{j=1}^{J} w_{dj}^2 \mathbb{V}(\mathbf g_{*j})+ \mathbb{V}(\mathbf h_{*})+\beta^{-1}I$, where $\mathbf y_*^\top=[\mathbf y_{*1}^\top,\dots,\mathbf y_{*D}^\top]$.
\subsection{Reconstruction}
For reconstruction, we compute the posterior density of $Y_*^m$ which is given below
\begin{eqnarray}
 \hspace{-7mm} &&p(Y_*^m|Y_*^{pt},Y)=\int  p(Y_*^m|\mathbf g_*^m,\mathbf h_*^m)p(\mathbf g_*^m,\mathbf h_*^m|X_*,Y_*^{pt},Y) p(X_*|Y_*^{pt},Y)d\mathbf h_*^m d\mathbf g_*^m dX_*.
\end{eqnarray}
Particularly, $p(X_*|Y_*^{pt},Y)$ is approximated by a Gaussian distribution $q(X_*)$ whose parameters need to be optimized to consider the partial observations $Y_*^{pt}$. This requires maximizing a new low bound of $\log p(Y,Y_*^{pt})$ which can be expressed as
\begin{align}
\widetilde{\mathcal{L}} = &\int q(X_*,X) \log p(Y_*^{pt},Y^{pt}|X_*,X)dX_*dX\\\nonumber
&+\int q(X)\log p(Y^m|X)dX - \mathrm{KL}[q(X_*,X)||p(X_*,X)].
\end{align}

This quantity $\widetilde{\mathcal{L}} $ can be maximized using the same method as training. In addition, parameters of the new variational distribution $q(X,X_*)$ are jointly optimized because $X$ and $X_*$ are coupled in $q(X,X_*)$.
\small
\bibliographystyle{IEEEtran}
\bibliography{report}

\end{document}